\documentclass[letterpaper, 10 pt, conference]{ieeeconf}  

\IEEEoverridecommandlockouts                              

\usepackage{graphics} 
\usepackage{epsfig} 
\usepackage{amsmath} 
\usepackage{amssymb}  
\usepackage{mathtools}
\usepackage{colortbl}
\usepackage{slashbox}
\usepackage{tabularx}
\usepackage{multirow}
\usepackage{url}
\usepackage{caption}
\usepackage{subcaption}
\usepackage[colorinlistoftodos,prependcaption,textsize=tiny]{todonotes}

\usepackage{algorithm}
\usepackage{algpseudocode}

\newcommand\kevin[1]{\textcolor{black}{#1}}

\newcommand{\insteadof}[1]{\ignorespaces}
\newcommand\martin[1]{\textcolor{black}{#1}}

\title{\LARGE \bf
Localising Faster: Efficient and precise lidar-based robot localisation in large-scale environments}
\author{Li Sun$^{1,3}$, Daniel Adolfsson$^{2}$, Martin Magnusson$^{2}$, Henrik Andreasson$^{2}$, Ingmar Posner$^{3}$, and Tom Duckett$^{1}$
  \thanks{$^{1}$Lincoln Centre for Autonomous Systems (L-CAS), University of Lincoln, UK
    {\tt\small \{lsun, tduckett\}@lincoln.ac.uk}}%
  \thanks{$^{2}$\"Orebro University, Sweden
    {\tt\small firstname.lastname@oru.se}}%
  \thanks{$^{3}$ University of Oxford
    {\tt\small \{kevin,ingmar\}@robots.ox.ac.uk}}%
}

\begin{document}

\maketitle
\thispagestyle{empty}
\pagestyle{empty}

\begin{abstract}
This paper proposes a novel approach for global localisation of mobile robots in large-scale environments. Our method leverages learning-based localisation and filtering-based localisation, to localise the robot efficiently and precisely through seeding Monte Carlo Localisation (MCL) with a deep-learned distribution. In particular, a fast localisation system rapidly estimates the 6-DOF pose through a deep-probabilistic model (Gaussian Process Regression with a deep kernel), then a precise recursive estimator refines the estimated robot pose according to the geometric alignment.
More importantly, the Gaussian method (i.e. deep probabilistic localisation) and non-Gaussian method (i.e. MCL) can be integrated naturally via importance sampling. Consequently, the two systems can be integrated seamlessly and mutually benefit from each other.
To verify the proposed framework, we provide a case study in large-scale localisation with a 3D lidar sensor. Our experiments on the Michigan NCLT long-term dataset show that the proposed method is able to localise the robot in 1.94 s on average (median of 0.8 s) with precision  0.75~m in a large-scale environment of approximately 0.5 km$^{2}$. 

\end{abstract}
\section{INTRODUCTION}
\label{sec:introduction}
For large-scale robotic applications in GPS-denied environments -- such as indoor industrial environments, underground mining, or space -- efficient and precise lidar-based robot localisation is in high demand. Geometry-based methods such as global registration \cite{ryde20103d,wolcott2015fast} and particle filters \cite{ueda2004expansion,yee2005grid,he2013observation,kucner-2015-mcl,bukhori2017detection} are widely used both to rescue a `kidnapped' robot and continuously localise the robot. However, the computational effort of these methods increases monotonically with the size of the environment. Deep learning methods \cite{kendall2015posenet,kendall2016modelling,kendall2017geometric,valada2018deep,radwan2018vlocnet++} are emerging in image-based relocalisation as a pivotal precursor to directly estimate the 6-DOF pose based on a model learned for a specific environment with a deep regression neural network. These learning methods are scalable as the computation time only depends on the complexity of the neural network. However, without geometric verification, they are likely to be less precise than geometry-based methods.

\begin{figure}[thpb]
  \centering
    \includegraphics[clip,trim=0mm 20mm 2mm 6mm, width=0.99\linewidth]{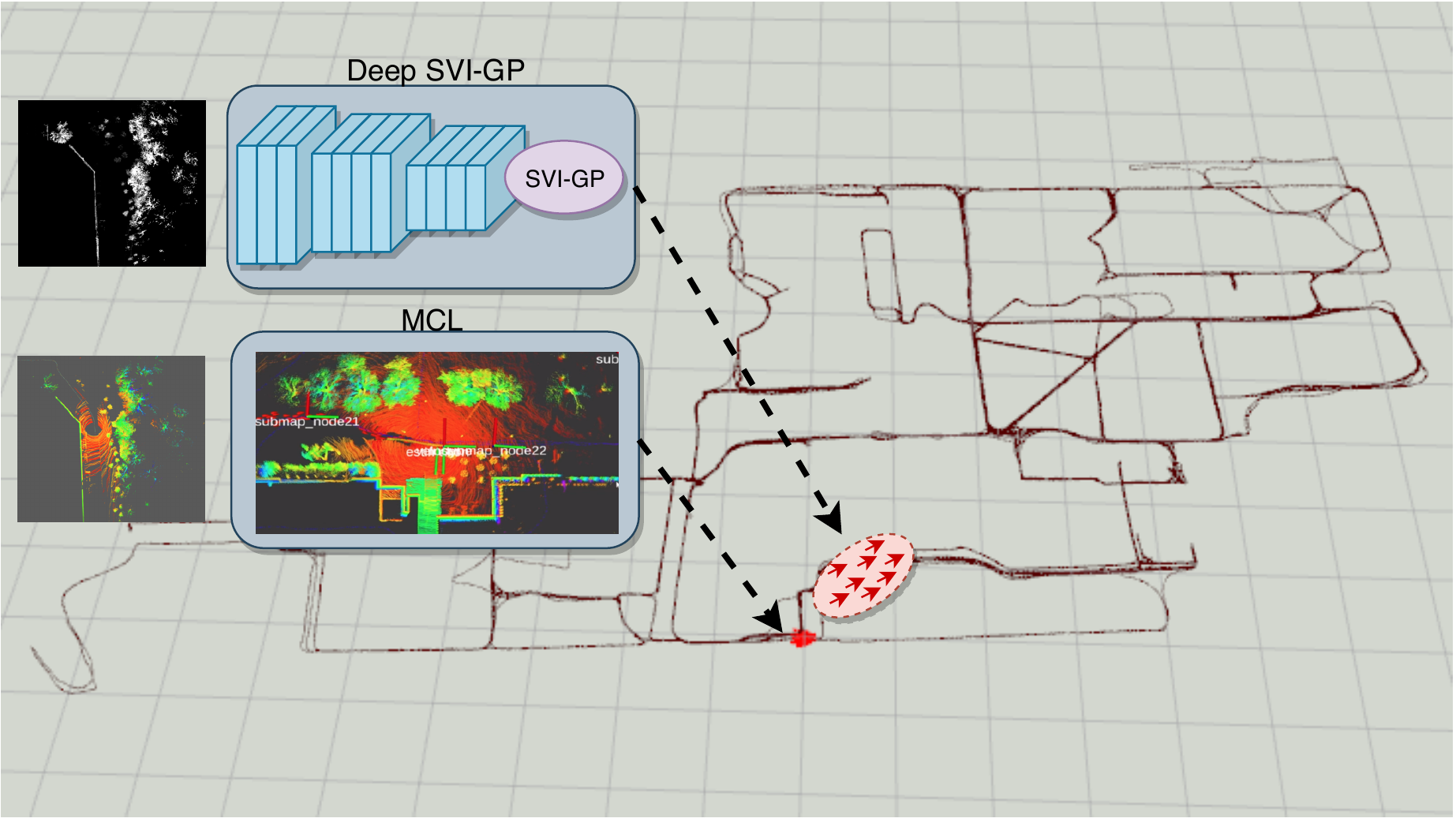}
   \caption{\label{fig:overall} We propose a hybrid global localisation method that enables precise localisation within 2 seconds on average on this large-scale map. Each cell in this grid is  50 $\times$ 50~m.}
   \vspace{-3ex}
\end{figure}

Using conventional filtering-based methods, e.g. Monte Carlo localisation (MCL), the pose estimate is updated recursively with each new observation. This tends to be very robust and lead to precise localisation, although when there is a large error in the initial estimate, it can take a long time for the filter to converge -- on the scale of minutes even for modest-sized maps~\cite{kucner-2015-mcl}. To mitigate this limitation, our intuition is to combine deep localisation and MCL. 

The first contribution of this paper is a hybrid localisation approach for lidar-based global localisation in large-scale environments. Our approach integrates 
a learning-based method with a Markov-filter-based method,
which makes it possible to efficiently provide global 
and continuous localisation with centimetre precision. 
A second contribution is a deep non-parametric model, i.e. a Gaussian Process with a conjunction of deep neural networks as the kernel,  used to estimate the distribution of the global 6-DOF pose from superimposed lidar scans. As a core component, the estimated distribution can be used to seed the Monte-Carlo localisation, thereby seamlessly integrating the two systems. A video demo is available at: {https://youtu.be/5wQXrpzxHNk}.
 
\section{RELATED WORK}
\label{sec:related_work}

\subsection{Lidar-based global localisation}
There exists a large body of research dealing with lidar-based localisation. 
The work that is most relevant to the scope of this paper can be generally divided into methods that provide informed initialisation of MCL, and appearance descriptors that aim to provide `one-shot' localisation.

While MCL in principle is robust and, by using a multimodal belief distribution, lends itself also to global localisation with 
a very uncertain initial estimate of the robot's location, a naive initialisation using a uniform particle distribution does not scale well to maps of realistic size. 
Some authors \cite{ryde20103d,yee2005grid}  alleviate the problem by working with a multi-resolution grid over the map. This makes it possible to scale to slightly larger maps, but is close to a uniform distribution and does not make use of the learned appearance of the map, as in our work.
Another strategy is to design a distribution from observations and sample particles from that \cite{thrun_monte_2000,he2013observation,kucner-2015-mcl}. 
Another possibility is to use external information, such as wi-fi beacons \cite{seow2017detecting} or a separate map layer that governs the prior probability of finding robots in each part of the map \cite{oh_map-based_2004}. 
In contrast, we work directly on point cloud data.
None of the above methods have been evaluated on maps as large
as those in our experiments.

Several engineered appearance descriptors have also been proposed for 2D \cite{tipaldi-2013-flirt}, 3D lidar-based global localisation \cite{magnusson2009automatic,schmiedel-2015-iron,cop2018delight}. Seg-Map \cite{dube2020segmap} \kevin{proposed to learn instance-level descriptors on point cloud segments for 3D lidar loop-closure detection}. 
These all have in common that they aim to provide `one-shot' global localisation, but  require a linear search over all descriptors created from the map. 
Even if the single descriptor look-up  is very quick, for large-scale long-term localisation, the linear scaling factor is still a major drawback compared to the method proposed in Sec.~\ref{sec:system1}.

\subsection{Deep image-based global localisation}

Image localisation is the task of accurately estimating the location and orientation of an image with respect to a global map, and has been studied extensively in both robotics and computer vision. 
Pose-Net \cite{kendall2015posenet} and related approaches \cite{kendall2016modelling,kendall2017geometric} have initiated a new trend to estimate the 6-DOF global pose using deep regression neural networks. In \cite{kendall2017geometric}, the geometric reprojection error, i.e. from the reprojection of the 3D reconstruction to the image frame, is jointly optimised with global pose loss during training. More recently, Valada et al.~\cite{valada2018deep} proposed geometry consistency loss to learn a spatially consistent global representation using relative pose loss as an auxiliary. Some researchers investigate the predictive uncertainties in the deep model \cite{kendall2016modelling,cai2018hybrid}, but sadly, uncertainties are not further utilised to improve the localisation. 


Global loop-closure detection can be combined with local feature matching as a hierarchical approach for visual relocalisation \cite{sarlin2018leveraging, Sarlin_2019_CVPR}. In these approaches, the deep global descriptors learned as a location signature are used to shortlist possible locations in a large-scale environment, then the precise 6-DOF pose can be estimated by 2D-to-3D matching between the retrieved key frames candidates and the map. With geometry verification, the localisation precision can be remarkably improved.

\section{METHODOLOGY}
\label{sec:methodology}


\begin{figure*}
  \centering
    \includegraphics[clip,trim=0mm 1.0mm 0mm 0.5mm,width=0.99\linewidth]{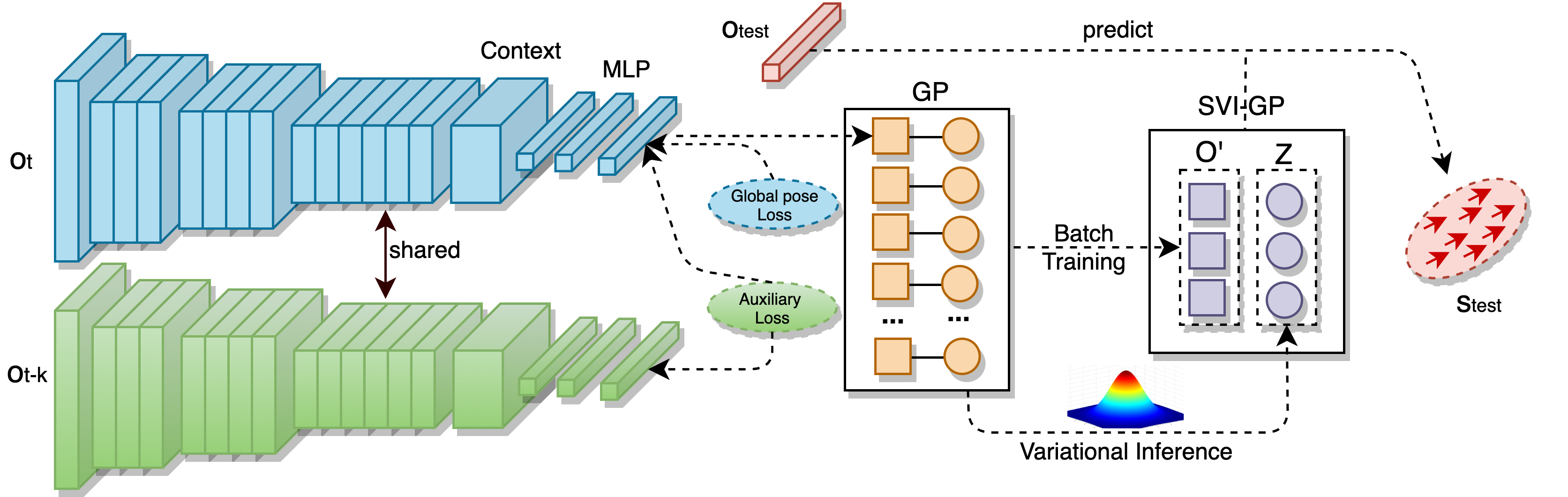}
   \caption{\label{fig:loc_net}The deep learning architecture. In the proposed method, we use the four convolutional blocks of ResNet-50 as backbone. A context stack consisting of 6 convolutional layers (128 filters with kernel size of 3$\times$3). and a triple layer MLP (i.e. 4096, 4096, 128) is used to learn the global location feature. Both global pose loss (consists of positional and rotational loss) and geometry-consistency loss are used to train the position and orientation features (using two branches). The final layer feature is used as the observation of GP to build the kernel. The SVI-GP is trained via a mini batch-training and latent variables are approximated by variational inference. All the network and GP parameters are trained end-to-end. }
  \vspace{-3ex}
\end{figure*}

\subsection{Problem Formulation}

In the global localisation problem, given the  observation $o_t$ and 6-DOF robot pose $s_t(p_t, r_t)$ at time $t$, the goal is to estimate the posterior $p(s_t|o_t)$. If the robot is moving, and a sequence of observations $o_{0:t}$ and control inputs (e.g. odometry) $u_{1:t}$ are available, the a-posteriori pose becomes $p(s_t|o_{0:t}, u_{1:t})$. 
\martin{Our proposed method has two systems working together for estimating this posterior.}

\emph{System~1} -- Efficient global localisation can be formulated through a learning-based method.
We aim to obtain a pose estimate from a single observation using a fast deterministic model. In order to train the deterministic model, the observations $O=\{o_i\}$ and poses $S=\{s_i\}$ used to build the map, i.e. $map=\mathrm{Mapping}(O, S)$, are provided as training examples. The problem can be formulated as estimating the conditional probability
$p(s_t|O, S, o_t)$. Parametric models (e.g. neural networks) or Gaussian methods (e.g. Gaussian Processes) can be used to resolve this.

\newcommand{\bel}{\mathrm{bel}} 

\emph{System~2} -- Precise localisation can be formulated 
through a Markov-filter-based method.
The a-posteriori belief state of the robot pose $\bel(s_t)$ can be iteratively updated as a Bayes filter as the robot is  moving. 
With the motion model $p(s_t|s_{t-1}, u_t)$, the belief can be updated as $\overline{\bel}(s_t) = \int p(s_t|s_{t-1}, u_t) \bel(s_{t-1}) \mathrm{d}s_{t-1}$ and given the measurement model $p(o_t|s_t)$, the belief can be then updated as $\bel(s_t) = \eta p(o_t|s_t) \overline{\bel}(s_t)$. 

To integrate the two systems, importance sampling can be used to update the particles generated by \emph{System 1}
using \emph{System 2}.
In particular, we propose to draw particles from 
$p(s_t|O, S, o_t)$ in \emph{System~1} and update belief ${\bel}(y_t)$ in \emph{System~2}, and maintain the particles in a healthy and converging distribution through importance sampling.

\subsection{\emph{System~1}: efficient localisation using a deep probabilistic model}\label{sec:system1}


To learn a deterministic model to efficiently predict the 6-DOF pose of the robot, a natural idea is to use a parametric statistical model like a deep neural network. A bonus is that the deep models can learn site-specific features through back-propagation. However, it is also important to model the predictive probability $p(s_t|O,S,o_t)$, which can efficiently generate robot pose hypotheses. Our intuition is to use a Gaussian process to estimate this conditional distribution.       

\subsubsection{Observation for learning} Using dense point clouds has a proven effectiveness in robot localisation \cite{cop2018delight}. To acquire a dense point cloud from a sparse lidar such as a Velodyne HDL-32E, without using extra mechanical devices, 
we first superimpose $k$ frames of consecutive observations  $\{o'_i\}_{i \in \{t-k,...,t\}}$ at time $t$ using odometry:
\begin{equation}
    o'_{t-k:t} = \Delta T_{t-k} o'_{t-k}  ... \cup \Delta T_{t-1} o'_{t-1} \cup o'_t
    ,
\end{equation}
where $\Delta T_{t-i}$ is the relative transformation between the pose at frame $t-i$ and that at frame $t$. $\Delta T_{t-i}$ can be obtained by a fusion of wheel odometry and inertial sensors, i.e. IMU and gyro, or lidar odometry. The superimposed point cloud can be converted to a height map and further encoded as a gray-scale bird's eye view image as shown in Fig. \ref{fig:overall}. 
In the learning of the deep probabilistic localisation, we use the bird's eye view image of the superimposed point cloud (denoted $o_{t} = o'_{t-k:t}$) as the observation at time $t$.

\subsubsection{Deep Neural Network for Feature Learning} A deep neural network is used to learn site-specific features from regressing 6-DOF global poses. Specifically, we first use five convolutional stacks of a pretrained ResNet-50 model as the backbone, and then the network is divided into two branches and triple-layer MLPs are used to learn the three-dimensional position $\hat{p} = (p_x, p_y, p_z)$ and four-dimensional rotation (quaternion) $\hat{r} = (q_x, q_y, q_z, q_w)$, respectively. We propose the following loss function for simultaneously minimising the positional loss and rotational loss.
\begin{equation}
    \mathcal{L}_{T,R} = ||\hat{p} - p^{gt}||_2 + \lambda ~(1-<\hat{r}, r^{gt}>^2) 
\end{equation}
Here, $<\hat{r}, r^{gt}>$ is the inner product of the predicted  and ground-truth quaternion. The second term indicates the distance between two normalised quaternion vectors.

Given a pair of bird's eye view images from time $t-k$ and $t$, the predicted global poses $\hat{T}_{t-k}$ and $\hat{T}_t$, and the ground truth poses $T_{t-k}^{gt}$ and $T_t^{gt}$, we can calculate the relative transform from the predictions and ground truth and make them geometrically consistent.
\begin{equation}
    \mathcal{L}_C = \mathcal{L}_{T,R} \big( ~(\hat{T}_{t-k})^{-1}\hat{T_{t}}, ~(T_{t-k}^{gt})^{-1}T_{t}^{gt}~ \big)
\end{equation}
More specifically, $\hat{T}_{t}$ ($T_T^{gt}$) is the transform matrix which can be obtained from the translation $\hat{p_t}$ ($p_t^{gt}$) and rotation $\hat{r_t}$ ($r_t^{gt}$) at time $t$. We convert the transform matrices back to pose vectors to compute the positional and rotational losses. We find that with the assistance of geometric consistency loss, the neural network can learn spatially consistent features and constrain the search space of global localisation, thereby enhancing the robustness of global pose estimation. 

\kevin{As mentioned in \cite{sattler2019understanding}, a learning-based model works well near training trajectories but might be challenging to use elsewhere in the mapped region. Hence we augment the training data in a region of 12.5~m to improve the performance.}\footnote{The bird's eye view image of size 400$\times$400 can be generated from the superimposed local point cloud in a visual scope of 100$\times$100$\times$10 metres with a resolution of 0.4 metres per pixel. We randomly crop a 300$\times$300 image for training and apply the corresponding translational offset to the target pose. For the geometry consistency learning, we randomly pick paired images within a window of ten frames ($k \in$[1,10]).}

The proposed deep neural network can learn site-specific and spatio-temporally consistent features. However, the inference (prediction) is not fully probabilistic with $L2$ loss. The drawback is that predictive uncertainties cannot be provided, i.e., the neural network cannot give the predictive distribution of the robot pose. In our two-system framework, the uncertainty of Bayesian localisation is of critical importance. An appropriate predictive distribution can accelerate the convergence of the MCL by giving small variances, and, at the same time, suppress the effects of false positives by predicting with large variances. To mitigate this, we adapt Gaussian process regression as the basis of the deep localisation network. That is, a hybrid probabilistic regression method is proposed where the deep neural network provides the deep kernel of a Gaussian Process. 

\subsubsection{Gaussian Process with Deep Kernel} Given the training observations $O$, learning target i.e. poses $S$, latent variable $f$, and the testing example $o_t$, the prediction step of Gaussian process regression involves inferring the conditional probability of the latent variable of testing example $p(f_*|O, S, o_*)$ as:  
\begin{equation}
\label{eq:gp_predict}
     \mathcal{N}(K_{*n}(K_{nn}+\sigma ^{2} I)^{-1} y, K_{**}- K_{*n}(K_{nn}+\sigma ^{2} I)^{-1} K_{n*})
\end{equation}

This conventional inference formula is not scalable as the computation increases exponentially with $n$, i.e. $\mathcal{O}(n^{3})$. Instead of using all the training examples for the prior $K_{nn}$, a reduced set of examples $O' \in \mathcal{R} ^{m*D}$ (known as the inducing points) is used to approximate the whole training set, where $D$ is the dimension of the feature and $m << n$. Given the latent variables of the inducing points, $Z=\{z_i\}$, the posterior distribution $p(Z|S)$ can be estimated by a variational distribution $Q(Z)$ (modelled as a multi-variant Gaussian, $Q(Z) \in \mathcal{N}(\mu, \Sigma)$). Via variational inference, the inducing points $O'$ can be estimated by maximization of the evidence lower bound (ELBO) of the log marginal likelihood of $p(S)$ \cite{titsias2009variational}:
\vspace{-4ex}

\begin{multline}
    \log p(S) \geq \mathcal{L}(Q(\mu, \Sigma), O')
    \\=\int Q(Z) \mathbb{E}_{p(f|Z)} [\log p(S|f)]\mathrm{d}Z - \mathit{KL} \big(Q(Z) || p(Z) \big).
    \vspace{-4ex}
\end{multline}
\vspace{-3ex}

In this formula, the first term is the predictive likelihood and \emph{KL} refers to the \emph{Kullback-Leibler} divergence. Titsias et al.~\cite{titsias2009variational} prove the final formula of the optimal inducing points $O'$ and the mean $\mu$ and variances $\Sigma$ of $Q(Z)$. In order to further make the training scalable, we train the SVIGP \cite{hensman2013gaussian} (Stochastic Variational Inference Gaussian Process) from mini-batch data.

With the optimised inducing points $O'$, and variational distribution $Q(Z)\in \mathcal{N}(\mu, \Sigma)$, the predictive probability in Eq. \ref{eq:gp_predict} can be reformulated as:
\vspace{-1ex}
\begin{equation}
\begin{split}
    p(f_*|O, S, o_*)=\int p(f_*|Z)Q(Z)\mathrm{d}Z =
    \mathcal{N}\big( f_*|K_{*m} K_{mm}^{-1} \mu, \\
    K_{**}-K_{*m}K_{mm}^{-1}K_{m*} + K_{*m}K_{mm}^{-1} \Sigma^{-1} K_{mm}^{-1} K_{m*} \big)~~~~~~
\end{split}
\label{eq:svigp}
\end{equation}
\vspace{-3ex}

We use a shared RBF kernel for multi-output prediction and the kernel is constructed from deep neural network features. To be more specific, the feature of the last layer (shown in Fig. \ref{fig:loc_net}) is used as the observation of the GP. By this means, the parametric neural network can be integrated with the non-parametric Gaussian Process.

More importantly, through maximising the log marginal likelihood, the inducing points $O'$, the variational distribution $Q(Z)$, hyper-parameters of the kernel $K$, and the parameters of the deep neural network can be jointly optimised by simple back-propagation-through-time. As the GP is very sensitive to the kernel parameters, to avoid suffering from local minima, we address the training in two stages. We first train the deep neural network, i.e. feature of the kernel, using the translational and rotational loss. Then, the GP is trained end-to-end with the deep kernels. 

\begin{figure}[h]
  \begin{subfigure}[b]{0.24\textwidth}
        \includegraphics[trim={9cm 9cm 15cm 0.5cm},clip,width=1.0\textwidth]{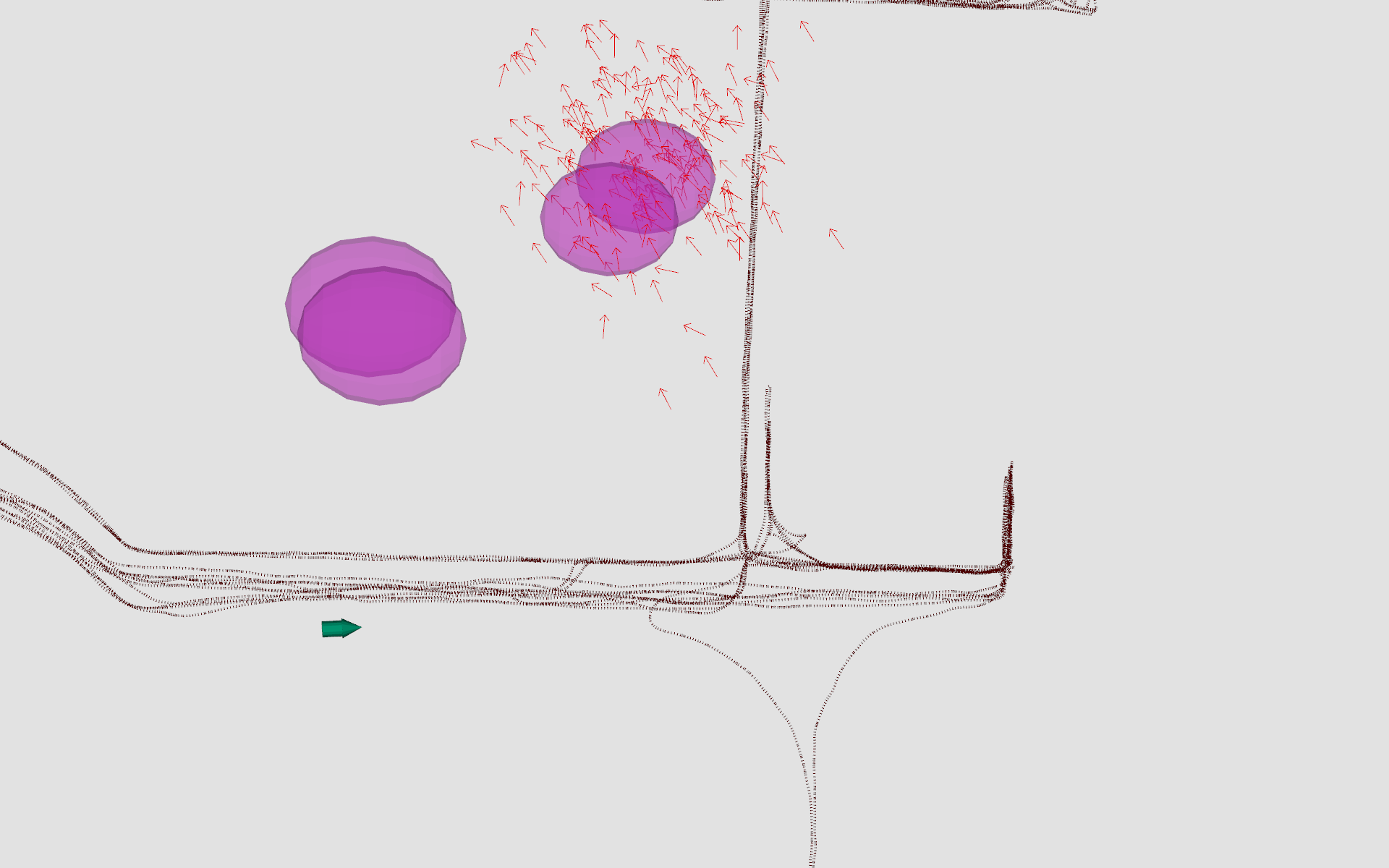}
        \caption{\raggedright\label{fig:typical_fails}
        Particles are sampled far from the robot position and diverge.
        }
  \end{subfigure}
    \begin{subfigure}[b]{0.237\textwidth}
        \includegraphics[trim={15cm 0cm 5cm 6cm},clip,width=1.0\textwidth]{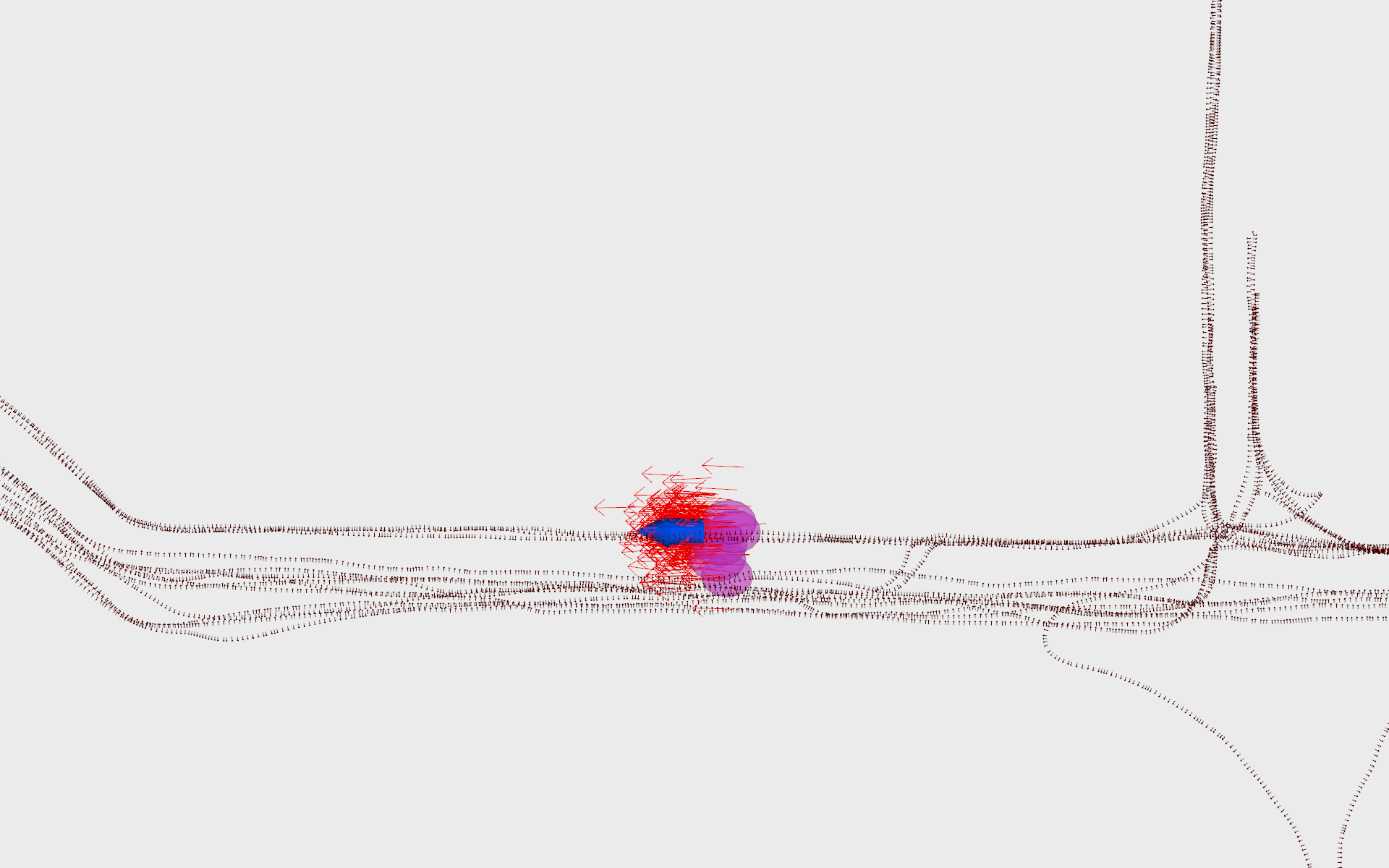}
        \caption{\raggedright\label{fig:typical_works}Particles are sampled close to the robot position with low uncertainty and converge.}
  \end{subfigure}
  \caption{\label{fig:gp_uncertainty_examples}{Qualitative results of success/failure cases.}
  Robot attempting to localise, true position is indicated by the green arrow. Pose particles(red arrows) are sampled according to the GP distribution(purple ellipsoids).
 }
  \vspace{-3ex}
\end{figure}

It is worth noting that we only train the Gaussian Process Regression for positioning, and the angular distance loss function is still used to learn the rotation.
This is for two reasons: firstly the inherent normalization attribute of the quaternion cannot be leveraged in the Gaussian Process via maximizing the log likelihood, and secondly, the predictive uncertainty of rotation is less important than that of position in large-scale localisation (in other words, rotational predictions depend on positional predictions). Qualitative results of the predictive distributions are shown in Fig. \ref{fig:gp_uncertainty_examples}.

\subsection{\emph{System~2}: Precise localisation using MCL}
For \emph{System 2} (MCL), 
we use a reference 3D map built using poses from RTK-GPS (for training) and we represent the map using the Normal Distribution Transform (NDT) for memory efficiency.
This step answers the $map=\mathrm{Mapping}(O,S)$ in the problem formulation. 

During localisation, with the motion model $p(s_t|s_{t-1}, u_t)$, the belief can be updated as
\begin{equation}
    \overline{\bel}(s_t) = \int p(s_t|s_{t-1}, u_t) \bel(s_{t-1}) \mathrm{d} s_{t-1}
    \vspace{-0.5ex}
\end{equation}
with the measurement model~\cite{thrun2005probabilistic}. This integral can be approximated via resampling with importance sampling, i.e. a particle filter. The a-posteriori belief estimate is updated as
\begin{equation}
    \bel(s_t) = \eta p(o_t|s_t) \overline{\bel}(s_t)
    ,
    \vspace{-0.5ex}
\end{equation}
where $\eta p(o_t|s_t)$ refers to the importance weights of samples, and the measurement model $p(o_t|s_t)$ can be obtained by calculating the distance between the lidar distribution and Map distribution with the 
Normal Distribution Transform (NDT) representation~\cite{saarinen2013normal}. 

Conventional methods usually first initialise a temporary particle distribution which is
reminiscent of the belief $\overline{\bel}(o_t)$. 
However, the belief can be estimated effectively from the current observation using the 
learning-based method,
hence 
providing 
a parametric method to initialise the belief as
\begin{equation}
\label{eq:integration}
    \overline{\bel}(s_t) = p(s_t|o_{1:t},u_{1:t}) \approx p(s_t|O, S, o_t)
    ,
\end{equation}
where the conditional probability $p(s_t|O, S, o_t)$ can be estimated by Gaussian Process $p(f_*|O, S, o_*)$ using Eq. \ref{eq:svigp}.

Practically, the particles are generated from two origins $\mathcal{S}_t = \mathcal{S}_{sys1} \cup \mathcal{S}_{sys2}$ in our implementation. They are the particles drawing from the GP's predictive distribution $\mathcal{S}_{sys1} \sim p(f_*|O, S, o_*)$ and particles $\mathcal{S}_{sys2}$ resampled from the previous belief set $\mathcal{S}_{t-1}$. Through the importance sampling mechanism, the $sys1$ particles from deep learning estimation and $sys2$ particles from MCL can be integrated, thereby integrating the two systems. 
A detailed description is shown in Algorithm \ref{alg:particle_filter}.
\vspace{-1ex}

\begin{algorithm}[h]
\begin{algorithmic}
\State \textbf{In:}
The map $map$, Gaussian Process model $GP$. Initially empty set ${S}_{t-1}$. Desired size $N_{sys1}$ and $N_{sys2}$. At each time stamp, the observation $o_t$, the control vector $u_t$.
\vspace{0.5mm}
\State \textbf{Out:}
The robot 6-DOF pose $s_t$.
\vspace{0.5mm}

\State
$\mathcal{S}_{sys2}$ = $\mathcal{S}_{t-1}$
\State
Draw $\mathcal{S}_{sys1}$ from $p(s_t|O,S,o_t)$ s.t. $|\mathcal{S}_{sys1}|=N_{sys1}$



\State
$\mathcal{S}_t = \mathcal{S}_{sys1} \cup \mathcal{S}_{sys2}$
\vspace{2mm}
\For{each particle $s_t^m$ in $\mathcal{S}_{t}$}
\State
Sample $s_t^m \sim p(s_t|s_{t-1}, u_t)$ 
\State
Update weight $w_t^m = p(o_t|s_t^m)$ 
\EndFor
\State
Normalise weights
\State
Resample $\mathcal{S}_t$ s.t. $|\mathcal{S}_t|=N_{sys2}$ according to weights

\vspace{2mm}
\State \Return 6-DOF pose $s_t = \frac{1}{N_{sys2}} \sum_m^{N_{sys2}} {w_t^ms_t^m}$.
\end{algorithmic}
\caption{A hybrid particle filtering approach}\label{alg:particle_filter}
\end{algorithm}
\vspace{-1ex}


\section{Experiment}

\begin{table*}[t]
\centering
\caption{Quantitative results of Bayesian localisation. }
\label{tab:exp_quantitative}
\begin{tabular}{l|lllllllll}
            \hline
             Metrics            &    Feb             & April            &     May        &  June           &    Aug           &    Oct            &    Nov             &   Dec             &    Overall \\             \hline
             median transitional error &   1.74m             &     1.69m       &   2.02m        &  1.99m          &    2.13m         &   2.14m           &    3.98m           &    3.59m           &  2.18m \\

             median rotational error   &     3.25$^\circ$   &    3.36$^\circ$  &  3.34$^\circ$  &  3.17$^\circ$     & 3.66$^\circ$   &   3.67$^\circ$   &   5.12$^\circ$      &    4.72$^\circ$    &  3.65$^\circ$\\ 
             
              
             mean transitional error &   8.77m           &    2.88m       &    15.3m     &   11.57m    &       14.06m        &    17.33m        &  30.15              &     32.08m   &  16.55m \\

             mean rotational error   &     6.19$^\circ$   &    4.43$^\circ$  &   9.50$^\circ$   &   7.96$^\circ$  &  8.52$^\circ$  &    11.58$^\circ$  &    17.98$^\circ$     &   14.51$^\circ$  &  9.99$^\circ$ \\
             
             \hline
             
             number of frames   &  33K     &      14K     &      26K        &    19K      &   27K      &       30K     &   14K       &   25K    & 184K \\  
  
                   \hline
\end{tabular}
\vspace{-0.5cm}
\end{table*}


Our research focuses on long-term localisation using 3D lidar data. In order to evaluate our proposed approach, a long-term mapping dataset with ground truth is required. \kevin{We did not choose the KITTI dataset~\cite{kitti} as the amount of overlapping trajectories are not sufficient for training}. To the best of our knowledge, the Michigan North Campus Long-Term Vision and lidar (NCLT) dataset~\cite{michigandataset} is the only long-term multi-session dataset currently available for lidar mapping and relocalisation. The dataset consists of 27 sessions with varying routes and days over 15 months. In each session, a Segway robot is driven via joystick to traverse the campus, and multi-sensor data including wheel odometry, 3D lidar, IMU, gyro, etc. are recorded. Ground-truth pose data are obtained by fusion of lidar scan matching and high precision RTK-GPS. The whole dataset spans 34.9~h  and 147.4~km. More details can be found in \cite{michigandataset}.  

Since the learning-based method requires training examples and the filter-based method needs a pre-built map, we selected eight sessions\footnote{Training sessions are 2012-01-08, 2012-01-15, 2012-01-22, 2012-02-02, 2012-02-04, 2012-02-05, 2012-03-31, and 2012-09-28.} for training and another eight sessions for testing. We selected the training sessions because they cover all explored areas of the campus, and the testing sessions were chosen randomly from varying seasons.   

Our hypothesis is that the learning-based method (\emph{System~1}) is efficient but lacks accuracy, while the filter-based method (\emph{System~2}) is precise but computationally intensive. By combining the two systems, efficient and precise localisation can be achieved.

\subsection{Deep Bayesian localisation evaluation}\label{sec:exp1}

\subsubsection{Training}
In this experiment, we evaluate the proposed deep probabilistic localisation and the learned uncertainties. The training of the 
neural network has two stages. Firstly, we train the network with $L2$ positional loss, angular orientation loss and auxiliary loss. Here the RES-Net stacks weights are transplanted from a pre-trained model (on ImageNet). In this stage, we use the Adam optimizer and train for 200 epochs with an initial learning rate of 10$^{-4}$ with exponential decay of 0.95. We clip the gradient by 5.0 and the learning rate by 10$^{-7}$. In the second stage, we use the same rotational loss and the last layer feature for position prediction to build the kernel of the Gaussian Process. We use 350 inducing points ($m$ in Eq. \ref{eq:svigp}) to estimate the variational distribution to approximate the prior for the whole dataset. More specifically, we transplant all the weights of the first stage to the deep GP model. We freeze the weights of the 5 ResNet stacks and optimise the parameters of the GP, fully-connected layers and context stack layers jointly by maximizing the log likelihood. In this stage, an initial learning rate of 10$^{-3}$ with exponential decay of 0.95 is used, and the model is trained for another 100 epochs. A discount of 0.1 and 0.01 on the learning rate is applied to the fully-connected layers and context stack layers. Our implementation is based on the TensorFlow
and GPflow\footnote{https://github.com/GPflow/GPflow} toolboxes. We use an i7 desktop with a NVIDIA TITAN X GPU for training, and the whole training process takes five days.

\subsubsection{Evaluation Criteria}
\label{sec:Criteria}

\begin{figure}[thpb]
\centering
\includegraphics[width=0.475\textwidth]{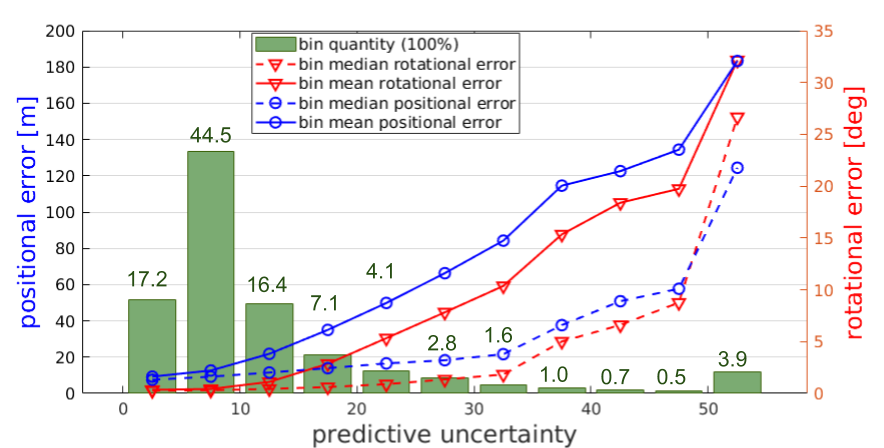}
\caption{\label{fig:global_localisation_exp}The Bayesian localisation performance under different uncertainty intervals. In this test, we uniformly divide the magnitude of uncertainties to 11 intervals from 0 to 50 (and above 50). The numbers above the bars indicate the percentages of testing examples in each bin.}
\vspace{-3ex}
\end{figure}



In this experiment, we follow the criteria used in \cite{kendall2015posenet} to evaluate the deep localisation. 
In particular, we use the positional error to measure the 3D position estimation, and the angular distance between predictive quaternion and ground truth quaternion to measure the rotational error. In this evaluation, the median errors are the more robust statistics for large-scale localisation. We evaluated our deep localisation model on eight testing sessions, and both the results on individual days and over the whole testing sets are provided in Table \ref{tab:exp_quantitative}.

As shown in Table \ref{tab:exp_quantitative}, we achieve an overall median positional and angular error of 2.18~m and 3.65$^\circ$ over eight sessions over the duration of one year. The median errors increase gradually from February to December from 1.74~m to 3.59~m, which can be attributed to the environment changes (i.e. plants and construction work) and new explored trajectories. Nevertheless, the reported performance shows satisfactory robustness to weather and seasonal variance, which demonstrates the capability of our 
model for long-term localisation.

\begin{table*}[t]
\centering
\begin{tabular}{ l|lllllllll|l|l }
 \hline
  &\multicolumn{9}{c}{Success rate (\%)}  &\multicolumn{2}{c} {Localisation time [s]} \\
 \hline
 \textbf{Method}       & Feb   & April  & May  & June   & Aug    & Oct   & Nov     & Dec  & Overall & Mean & Median\\
 \hline
 Hybrid - GP Cov (ours)  &  $97.2$  & $\bf 100$   & $\bf 94.4$  & $\bf 94.9$  &  $\bf 95.8$  & $\bf 94.7$  & $\bf 81.2$    & $\bf 88.7$ & $\bf 93.3$  & $\bf 1.94  \pm 3.0$  & $\bf 0.80$\\
  Hybrid - Fixed Cov (ours) & $\bf 97.7$   & $99.0$    & $93.0$  & $93.5$    & $94.8$   & $94.0$    & $79.7$    & $88.3$ & $92.5$ & $2.32 \pm 3.3$ & $0.95$\\
 NDT-MCL with uniform initialisation  & $62.0$    & $70.6$  & $57.3$   &  $59.6$   & $52.7$  & $51.7$    & $37.6$ & $40.1$ & $54.0$ & $154.29\pm 46.2$ & $157.93$  \\
 \hline
\end{tabular}
\caption{Success rate of hybrid localisation with fixed and GP covariance compared to Uniform MCL.}\label{tab:success}
\vspace{-3ex}
\end{table*}

\subsubsection{Uncertainty Evaluation}
We further evaluate the predictive probabilities of our proposed 
model. 
It is more important to predict locations with uncertainties, which is the advantage of non-parametric models, e.g. Gaussian Process, compared to parametric deep neural networks. In the 
hybrid method,
a well modelled predictive probability distribution is able to accelerate the convergence of the particle filter, but can also suppress ill-posed localisation due to false positives. 

To evaluate the uncertainties, we divide the magnitude of the predictive uncertainties ($L2$ norm of variances in x, y, and z directions). We divide the magnitude of uncertainties (i.e. of positional prediction) into uniform intervals and calculate the mean and median errors within 
intervals. The histogram of predictions is also counted. The statistical results are shown in Fig. \ref{fig:global_localisation_exp}. We found that both positional error and rotational error positively propagate to the magnitude of uncertainties. Most of the predictions ($\geq$85\%) fall into the first four bins, i.e. the magnitude of uncertainty is less than 20. Within this uncertainty range (0--20 in magnitude), the proposed model achieves positional errors of less than 4.3~m on average and 2.0~m median, and rotational errors of less than 2.7$^\circ$ on average and 1.7$^\circ$ median. 

\subsection{Monte-Carlo Localisation Baseline Evaluation}\label{sec:exp2}


For large-scale environments like the Michigan campus, the large amount of particles required for naive MCL initialisation makes MCL intractable. For that reason, we restricted sampling only around previously explored positions from the training dataset.
 Specifically, for each explored position, 3$\times$3$\times$3 points were sampled from a voxel grid with voxel size $v_x$, $v_y$, $v_z$ of $0.2$~m. A finer resolution would make the amount of particles unmanageable.
 Nearby points were then filtered using a voxel-filter with the same resolution. For each remaining point, 8 pose particles were created with evenly spaced orientations around the z-axis.
During the resampling step, 
the number of particles was reduced by a fraction of $0.6$ until $1000$ particles remained. 

In total, over 4000 localisation attempts (initial locations are uniformly chosen from 8 sessions) were performed using our two methods and the MCL baseline. 
An attempt was considered successful if an error $<0.75$~m was achieved within $140$ iterations. MCL scored a $54\%$ success rate with an average localisation time of $154.3$~s as shown in Table~\ref{tab:success}.

\subsection{Hybrid Localisation Evaluation}\label{sec:exp3}

\begin{figure}
  \centering
  \begin{subfigure}[b]{0.5\textwidth}
  \centering
        \includegraphics[clip, trim=0.2cm 0.75cm 0cm 0cm,width=0.99\textwidth]{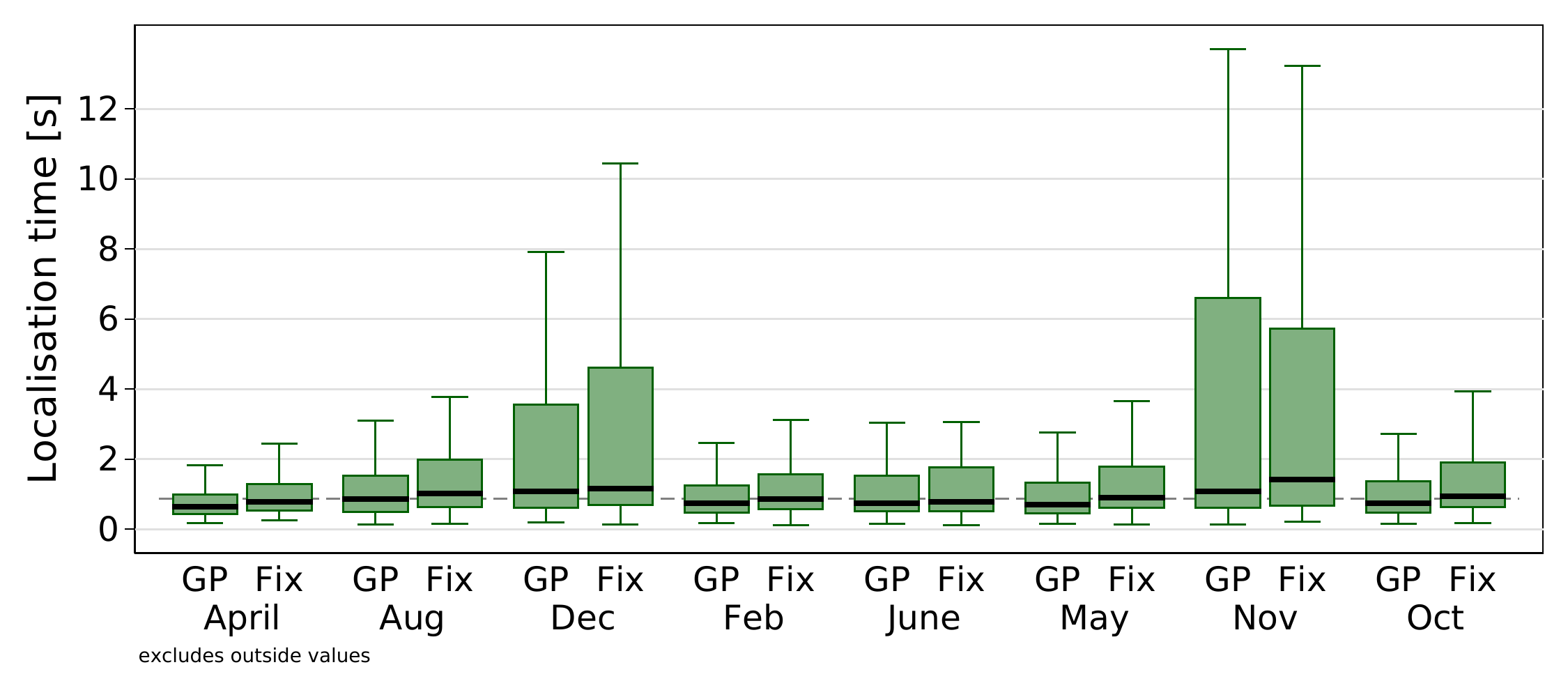}
  \end{subfigure}
\caption{Localisation time using our method with fixed covariance (Fix) and covariance from Gaussian Process (GP). Localisation time is higher in November and December which follows from the higher failure rate.
}\label{fig:hybrid_localisation_time}
\vspace{-3ex}
\end{figure}

Instead of initialising particles in all possible locations, we use the proposed deep probabilistic method to seed the MCL (Eq. \ref{eq:integration}) and continue to update the samples following Algorithm \ref{alg:particle_filter}. 
Specifically, we used a nominal number of 500 particles, increasing up to 1000 as additional particles were sampled, and reducing back to 500 during the resampling step.
Using only a small amount of particles with the fast and sparse NDT-based measurement likelihood model, we achieved an average iteration time of 0.073~s with $\sigma=$0.02. 
To investigate the benefit of sampling from the uncertainty estimates, we compared it to sampling from a fixed position distribution ($\sigma_x^2=70, \sigma_y^2=70, \sigma_z^2=3$). Orientations were sampled from a fixed distribution($\sigma_{ex}^2=0.0225, \sigma_{ey}^2=0.25, \sigma_{ez}^2=0.0225$) in both cases. These parameters were chosen according to practical experience. The localisation success rate and speed is shown in table.~\ref{tab:success} and Fig.~\ref{fig:hybrid_localisation_time}.

We found a median and mean localisation time of 0.799~s and 1.944~s respectively for the hybrid approach with covariance estimated by GP. Compared to the MCL baseline, the median localisation time is reduced by $99.5\%$.
Similarly to the evaluation in Sec.~\ref{sec:Criteria}, the highest success rate was obtained in the early months of the year, gradually decreasing over the year. \kevin{We observed in November and December, the localisation time significantly increased because there are more novel trajectories and landmarks.}


\section{Conclusion}
This paper proposes a hybrid probabilistic localisation method which leverages the efficient inference of the deep deterministic model and the rigorous geometry verification of Bayes-filter-based localisation. This paper seeks a solution to resolve the non-conjugacy between the Gaussian method (Gaussian Process) and non-Gaussian method (Monte-Carlo localisation) through importance sampling. Consequently, the two systems can be integrated seamlessly.

From the experiments, we found that the learning-based localisation method can provide an optimised predictive distribution to seed MCL, thereby accelerating the convergence of particles. On the other hand, the false positives can be suppressed by the correctly modelled uncertainties in the continuous localisation. The experimental results show that the hybrid system is able to localise in $99.5\%$ less time compared to the Monte-Carlo baseline method, i.e. NDT-MCL, and increases the precision to centimetres to meet the needs of large-scale real-world localisation problems.

\section*{ACKNOWLEDGMENT}
We thank NVIDIA Co. for donating a high-power GPU on which this work was performed. This project has received funding from the EU Horizon 2020 under grant agreement No 732737 (ILIAD) and EPSRC Programme Grant EP/M019918/1.

\bibliographystyle{IEEEtran}
{\bibliography{refs}}

\begin{thebibliography}{10}
\providecommand{\url}[1]{#1}
\csname url@rmstyle\endcsname
\providecommand{\newblock}{\relax}
\providecommand{\bibinfo}[2]{#2}
\providecommand\BIBentrySTDinterwordspacing{\spaceskip=0pt\relax}
\providecommand\BIBentryALTinterwordstretchfactor{4}
\providecommand\BIBentryALTinterwordspacing{\spaceskip=\fontdimen2\font plus
\BIBentryALTinterwordstretchfactor\fontdimen3\font minus
  \fontdimen4\font\relax}
\providecommand\BIBforeignlanguage[2]{{%
\expandafter\ifx\csname l@#1\endcsname\relax
\typeout{** WARNING: IEEEtran.bst: No hyphenation pattern has been}%
\typeout{** loaded for the language `#1'. Using the pattern for}%
\typeout{** the default language instead.}%
\else
\language=\csname l@#1\endcsname
\fi
#2}}

\bibitem{ryde20103d}
J.~Ryde and H.~Hu, ``3d mapping with multi-resolution occupied voxel lists,''
  \emph{Autonomous Robots}, vol.~28, no.~2, p. 169, 2010.

\bibitem{wolcott2015fast}
R.~W. Wolcott and R.~M. Eustice, ``Fast lidar localization using
  multiresolution gaussian mixture maps,'' in \emph{2015 IEEE international
  conference on robotics and automation (ICRA)}.\hskip 1em plus 0.5em minus
  0.4em\relax IEEE, 2015, pp. 2814--2821.

\bibitem{ueda2004expansion}
R.~Ueda, T.~Arai, K.~Sakamoto, T.~Kikuchi, and S.~Kamiya, ``Expansion resetting
  for recovery from fatal error in {Monte Carlo} localization-comparison with
  sensor resetting methods,'' in \emph{2004 IEEE/RSJ International Conference
  on Intelligent Robots and Systems (IROS)}, vol.~3.\hskip 1em plus 0.5em minus
  0.4em\relax IEEE, 2004, pp. 2481--2486.

\bibitem{yee2005grid}
M.~Y. Yee and J.~Vermaak, ``A grid-based proposal for efficient global
  localisation of mobile robots,'' in \emph{Proceedings.(ICASSP'05). IEEE
  International Conference on Acoustics, Speech, and Signal Processing, 2005.},
  vol.~5.\hskip 1em plus 0.5em minus 0.4em\relax IEEE, 2005, pp. v--217.

\bibitem{he2013observation}
T.~He and S.~Hirose, ``Observation-driven bayesian filtering for global
  location estimation in the field area,'' \emph{Journal of Field Robotics},
  vol.~30, no.~4, pp. 489--518, 2013.

\bibitem{kucner-2015-mcl}
T.~P. Kucner, M.~Magnusson, and A.~J. Lilienthal, ``Where am {I}?: An
  {NDT}-based prior for {MCL},'' in \emph{Proceedings of the European
  Conference on Mobile Robots (ECMR)}, Sept. 2015.

\bibitem{bukhori2017detection}
I.~Bukhori and Z.~H. Ismail, ``Detection of kidnapped robot problem in {Monte
  Carlo} localization based on the natural displacement of the robot,''
  \emph{International Journal of Advanced Robotic Systems}, vol.~14, no.~4, p.
  1729881417717469, 2017.

\bibitem{kendall2015posenet}
A.~Kendall, M.~Grimes, and R.~Cipolla, ``Posenet: A convolutional network for
  real-time 6-dof camera relocalization,'' in \emph{Proceedings of the IEEE
  international conference on computer vision}, 2015, pp. 2938--2946.

\bibitem{kendall2016modelling}
A.~Kendall and R.~Cipolla, ``Modelling uncertainty in deep learning for camera
  relocalization,'' in \emph{2016 IEEE international conference on Robotics and
  Automation (ICRA)}.\hskip 1em plus 0.5em minus 0.4em\relax IEEE, 2016, pp.
  4762--4769.

\bibitem{kendall2017geometric}
------, ``Geometric loss functions for camera pose regression with deep
  learning,'' in \emph{Proceedings of the IEEE Conference on Computer Vision
  and Pattern Recognition}, 2017, pp. 5974--5983.

\bibitem{valada2018deep}
A.~Valada, N.~Radwan, and W.~Burgard, ``Deep auxiliary learning for visual
  localization and odometry,'' in \emph{2018 IEEE International Conference on
  Robotics and Automation (ICRA)}.\hskip 1em plus 0.5em minus 0.4em\relax IEEE,
  2018, pp. 6939--6946.

\bibitem{radwan2018vlocnet++}
N.~Radwan, A.~Valada, and W.~Burgard, ``Vlocnet++: Deep multitask learning for
  semantic visual localization and odometry,'' \emph{IEEE Robotics and
  Automation Letters}, vol.~3, no.~4, pp. 4407--4414, 2018.

\bibitem{thrun_monte_2000}
S.~Thrun, D.~Fox, W.~Burgard, and {others}, ``{Monte Carlo }localization with
  mixture proposal distribution,'' in \emph{{AAAI}/{IAAI}}, 2000, pp. 859--865.

\bibitem{seow2017detecting}
Y.~Seow, R.~Miyagusuku, A.~Yamashita, and H.~Asama, ``Detecting and solving the
  kidnapped robot problem using laser range finder and wifi signal,'' in
  \emph{2017 IEEE international conference on real-time computing and robotics
  (RCAR)}.\hskip 1em plus 0.5em minus 0.4em\relax IEEE, 2017, pp. 303--308.

\bibitem{oh_map-based_2004}
S.~M. Oh, S.~Tariq, B.~N. Walker, and F.~Dellaert, ``Map-based priors for
  localization,'' in \emph{{IEEE}/{RSJ} {International} {Conference} on
  {Intelligent} {Robots} and {Systems}({IROS}).}, vol.~3.\hskip 1em plus 0.5em
  minus 0.4em\relax IEEE, 2004, pp. 2179--2184.

\bibitem{tipaldi-2013-flirt}
G.~D. {Tipaldi}, L.~{Spinello}, and W.~{Burgard}, ``Geometrical {FLIRT} phrases
  for large scale place recognition in 2d range data,'' in \emph{2013 IEEE
  International Conference on Robotics and Automation}, May 2013, pp.
  2693--2698.

\bibitem{magnusson2009automatic}
M.~Magnusson, H.~Andreasson, A.~N{\"u}chter, and A.~J. Lilienthal, ``Automatic
  appearance-based loop detection from three-dimensional laser data using the
  normal distributions transform,'' \emph{Journal of Field Robotics}, vol.~26,
  no. 11-12, pp. 892--914, 2009.

\bibitem{schmiedel-2015-iron}
T.~Schmiedel, E.~Einhorn, and H.-M. Gross, ``{IRON}: A fast interest point
  descriptor for robust {NDT}-map matching and its application to robot
  localization,'' in \emph{IEEE/RSJ International Conference on Intelligent
  Robots and Systems}, 2015.

\bibitem{cop2018delight}
K.~P. Cop, P.~V. Borges, and R.~Dub{\'e}, ``Delight: An efficient descriptor
  for global localisation using lidar intensities,'' in \emph{2018 IEEE
  International Conference on Robotics and Automation (ICRA)}.\hskip 1em plus
  0.5em minus 0.4em\relax IEEE, 2018, pp. 3653--3660.

\bibitem{dube2020segmap}
R.~Dub{\'e}, A.~Cramariuc, D.~Dugas, H.~Sommer, M.~Dymczyk, J.~Nieto,
  R.~Siegwart, and C.~Cadena, ``Segmap: Segment-based mapping and localization
  using data-driven descriptors,'' \emph{The International Journal of Robotics
  Research}, vol.~39, no. 2-3, pp. 339--355, 2020.

\bibitem{cai2018hybrid}
M.~Cai, C.~Shen, and I.~D. Reid, ``A hybrid probabilistic model for camera
  relocalization.'' in \emph{BMVC}, vol.~1, no.~2, 2018, p.~8.

\bibitem{sarlin2018leveraging}
P.-E. Sarlin, F.~Debraine, M.~Dymczyk, and R.~Siegwart, ``Leveraging deep
  visual descriptors for hierarchical efficient localization,'' in
  \emph{Conference on Robot Learning}, 2018, pp. 456--465.

\bibitem{Sarlin_2019_CVPR}
P.-E. Sarlin, C.~Cadena, R.~Siegwart, and M.~Dymczyk, ``From coarse to fine:
  Robust hierarchical localization at large scale,'' in \emph{The IEEE
  Conference on Computer Vision and Pattern Recognition (CVPR)}, June 2019.

\bibitem{sattler2019understanding}
T.~Sattler, Q.~Zhou, M.~Pollefeys, and L.~Leal-Taixe, ``Understanding the
  limitations of cnn-based absolute camera pose regression,'' in
  \emph{Proceedings of the IEEE Conference on Computer Vision and Pattern
  Recognition}, 2019, pp. 3302--3312.

\bibitem{titsias2009variational}
M.~Titsias, ``Variational learning of inducing variables in sparse {Gaussian}
  processes,'' in \emph{Artificial Intelligence and Statistics}, 2009, pp.
  567--574.

\bibitem{hensman2013gaussian}
J.~Hensman, N.~Fusi, and N.~D. Lawrence, ``Gaussian processes for big data,''
  in \emph{Proceedings of the Twenty-Ninth Conference on Uncertainty in
  Artificial Intelligence}.\hskip 1em plus 0.5em minus 0.4em\relax AUAI Press,
  2013, pp. 282--290.

\bibitem{thrun2005probabilistic}
S.~Thrun, W.~Burgard, and D.~Fox, \emph{Probabilistic robotics}.\hskip 1em plus
  0.5em minus 0.4em\relax MIT press, 2005.

\bibitem{saarinen2013normal}
J.~Saarinen, H.~Andreasson, T.~Stoyanov, and A.~J. Lilienthal, ``Normal
  distributions transform monte-carlo localization (ndt-mcl),'' in \emph{2013
  IEEE/RSJ International Conference on Intelligent Robots and Systems}.\hskip
  1em plus 0.5em minus 0.4em\relax IEEE, 2013, pp. 382--389.

\bibitem{kitti}
A.~Geiger, P.~Lenz, and R.~Urtasun, ``Are we ready for autonomous driving? the
  kitti vision benchmark suite,'' in \emph{Conference on Computer Vision and
  Pattern Recognition (CVPR)}, 2012.

\bibitem{michigandataset}
N.~Carlevaris-Bianco, A.~K. Ushani, and R.~M. Eustice, ``University of
  {Michigan} {North} {Campus} long-term vision and lidar dataset,''
  \emph{International Journal of Robotics Research}, vol.~35, no.~9, pp.
  1023--1035, 2015.

\end{thebibliography}

\end{document}